\documentclass[conference]{IEEEtran}
\usepackage{hyperref}
\usepackage{cite}
\usepackage{amsmath,amssymb,amsfonts}
\usepackage{algorithmic}
\usepackage{graphicx}
\usepackage{textcomp}
\usepackage{listings}
\usepackage{verbatim}
\def\BibTeX{{\rm B\kern-.05em{\sc i\kern-.025em b}\kern-.08em
    T\kern-.1667em\lower.7ex\hbox{E}\kern-.125emX}}

\usepackage[caption=false]{subfig}
\usepackage{adjustbox}
\usepackage{fancyvrb}

\usepackage{enumitem}

\usepackage[table]{xcolor}
\definecolor{lightgray}{gray}{0.9}

\lstdefinestyle{customPython}{
  language=Python,
  tabsize=4,
  showspaces=false,
  showstringspaces=false,
  numbersep=10pt,
}

\usepackage{floatrow}
\newfloatcommand{capbtabbox}{table}[][\FBwidth]

\DeclareMathOperator*{\argmax}{argmax}

\begin{document}

\setlength{\abovedisplayskip}{3pt}
\setlength{\belowdisplayskip}{3pt}
\setlength{\abovedisplayshortskip}{3pt}
\setlength{\belowdisplayshortskip}{3pt}

\title{Robotic Surgery With Lean Reinforcement Learning}

\author{
  \IEEEauthorblockN{Yotam Barnoy\IEEEauthorrefmark{1},
  Molly O'Brien\IEEEauthorrefmark{2},
  Will Wang\IEEEauthorrefmark{3},
  Gregory Hager\IEEEauthorrefmark{4}}

  \IEEEauthorblockA{Department of Computer Science,
  The Johns Hopkins University\\
  Email: \IEEEauthorrefmark{1}ybarnoy1@cs.jhu.edu,
  \IEEEauthorrefmark{2}mobrie38@jhu.edu,
  \IEEEauthorrefmark{3}wwang121@jhu.edu,
  \IEEEauthorrefmark{4}hager@cs.jhu.edu}
}

\maketitle

\footnotetext{
The dVSS code is closed source, but our RL framework,
together with the Needle Master environment,
is available at
\url{https://github.com/bluddy/lean_rl}
}


\begin{abstract}

As surgical robots become more common,
automating away some of the burden of complex direct human operation becomes ever more feasible.
Model-free reinforcement learning (RL) is a promising direction toward
generalizable automated surgical performance,
but progress has been slowed by the lack of efficient and realistic learning environments.
In this paper, we describe adding reinforcement learning support to the da Vinci Skill Simulator,
a training simulation used around the world to allow surgeons to learn and rehearse technical skills.
We successfully teach an RL-based agent to perform sub-tasks in the simulator environment,
using either image or state data.
As far as we know, this is the first time an RL-based agent is taught from visual data in a surgical robotics environment.
Additionally, we tackle the sample inefficiency of RL
using a simple-to-implement system which we term hybrid-batch learning (HBL),
effectively adding a second, long-term replay buffer to the Q-learning process.
Additionally, this allows us to bootstrap learning from images
from the data collected using the easier task of learning from state.
We show that HBL decreases our learning times significantly.
\end{abstract}

\begin{IEEEkeywords}
Surgery, Robotics, Reinforcement Learning
\end{IEEEkeywords}


\section{Introduction}

Robotic surgery is becoming ever more popular due to such innovations as the da Vinci\cite{davinci2000}
robot.
Currently, such a robot is directly driven by a human operator sitting at the control console,
manipulating the Master Tool Manipulator (MTM),
while a computer translates the movement to the Patient Side Manipulators (PSM).
Given the fact that a computer moderates the input,
it would be convenient to allow a surgeon to delegate simple tasks such as wound suturing to the computer,
permitting the surgeon to focus on more high-level tasks.

We investigate this automation task in the context of a simulation provided to us by Intuitive Surgical
called the da Vinci Skill Simulator (dVSS).
This simulation replaces the physical robotic arms while allowing the surgeon to use the same Surgeon's Console.
The dVSS is used by surgeons to rehearse technical skills for robotic surgery~\cite{dvss2019}.

Robotic surgery involves high stakes and complex environments.
While gradual progress is being made to automate parts of surgery~\cite{dettorre_2018},~\cite{sundaresan_2019},
they require tailor-made `classical' solutions to individual sub-tasks.
Reinforcement learning(RL), on the other hand, offers the potential benefit of being applicable
to multiple environmental conditions and tasks without requiring a human to generate specific solutions
once conditions change.
Furthermore, using visual input in an RL system allows the system to consult a succint summary of relevant state that is not
easily encapsulated or normally considered relevant.

We focus on model-free RL,
which has had many recent successes and requires relatively low per-task customization, to automate our tasks.
Deep RL was first applied to Atari game environments~\cite{mnih2013playing},
but it has recently also been successfully applied to robotics~\cite{Zeng2018}\cite{Hundt2019}.
However, robotic surgery presents unique challenges to RL
while most recent robotic tasks involve rigid bodies being moved around by robotic arms,
robotic surgery involves robotic arms manipulating soft tissue.
Additionally, surgical tasks have a low error tolerance and a small risk of catastrophic error,
and teaching RL agents on a real robot introduces potential costs and slow task setup times.

\begin{figure}[t!]%

  \centering
  \subfloat[Reach task view]{\includegraphics[width=0.48\linewidth,trim=0 0 0 40,clip]{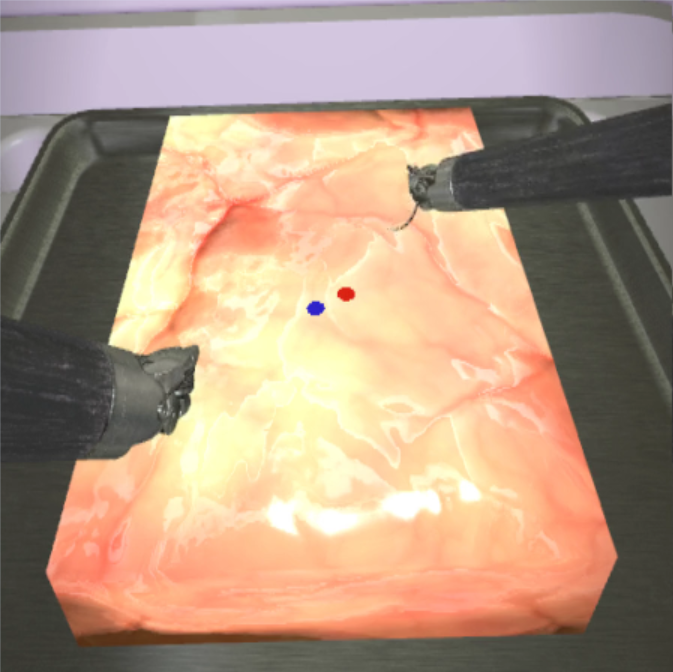}}%
  \hspace{0.02\linewidth}
  \subfloat[Suture task view]{\includegraphics[width=0.48\linewidth,trim=0 10 0 40,clip]{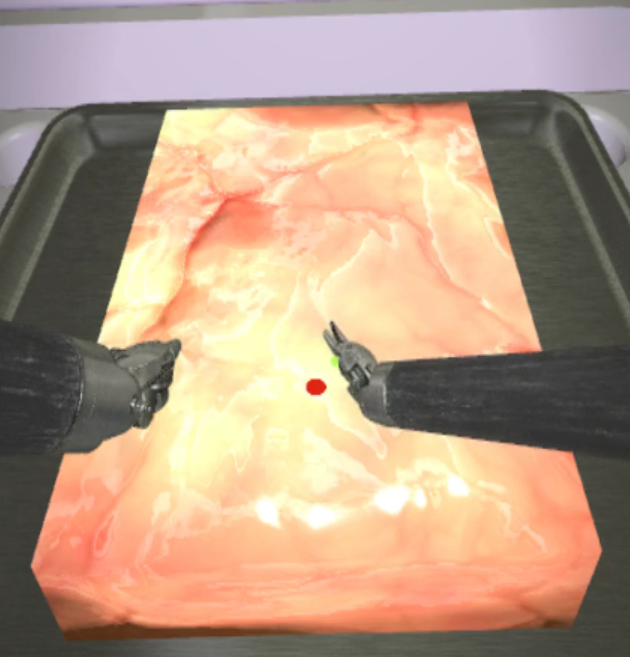}}%

\caption{%
  The environmental view from the perspective of the user/agent,
  showing the tissue, the PSM (Patient-Side Manipulator) arms of the da Vinci robot, and the needle.
  On the left, we see the view of the reach task,
  where the gripper needs to reach the location of the blue target.
  On the right, we see the suture task,
  where the agents needs to manipulate the needle so that it enters at the ingress
  position (normally in blue, but turns green post-ingress),
  and exits at the egress position (in red).
  }
\label{fig:view}
\end{figure}

We address these issues by using the dVSS,
and modify the simulation to form the dVSS-RL: an RL environment (see Figure~\ref{fig:view})
which utilizes the kinematics of the da Vinci robot;
the rigid dynamics of the da Vinci arms and needle;
and the soft-body dynamics of the needle interacting with tissue.
This allows for millions of learning iterations without the need for a real robot.

Training with the dVSS-RL requires on the order of millions of steps due to the fact that RL
is fairly sample-inefficient~\cite{mnih2013playing}.
One approach to this problem is to run experiments on large-scale compute resources~\cite{vinyals2019alphastar}.
Recognizing this as a common problem with RL in both robotics and complex real-time environments,
we choose a multi-pronged `lean' approach instead:
first, inspired by~\cite{Zeng2018} and~\cite{Hundt2019},
we reduce the complexity of the RL problem within reason,
pruning out unnecessary dimensions.
Second, we run multiple copies of the dVSS-RL in parallel, allowing for faster progress.
Third, we utilize a scheme which we term Hybrid Batch Learning, that effectively adds a long-term replay buffer
in addition to the normal replay buffer found in Q-learning.
This allows us to learn simultaneously from a mix of off-policy and on-policy data,
and to reuse the results of failed early runs to
create more successful runs later on.
Using this method, we additionally leverage a form of curriculum learning~\cite{Bengio2009},
progressing from easier problems such as learning to manipulate the dVSS-RL directly from state,
to harder ones, such as learning to do so from images.

Our main contributions are as follows:

\begin{itemize}

  \item
    We apply both state- and image-based model-free RL to learn a suturing task
    utilizing the dVSS - an official surgical simulation used by surgeons to rehearse technical skills.
    As far as we know, our algorithm is the first surgical model-free RL algorithm to work based on vision,
    and includes needle-tissue interaction.

  \item
    We apply a second, long-term replay buffer, which we term Hybrid-Batch Learning.
    to accelerate this task.
    We replay batch off-policy data while simultaneously learning in on-policy environments.
    This scheme allows us to learn in far less time than using purely on-policy environments,
    allowing for around 6x improvement in performance.
    We also use this scheme to apply a form of curriculum learning,
    using the same batch data from our state learning to accelerate the image-based learning.

\end{itemize}


\section{Related Work}

\subsection{Robotic Reinforcement Learning}

RL has been applied to many environments, including
Atari games~\cite{mnih2013playing},
autonomous driving frameworks~\cite{el2017deep}\cite{Kendall},
and first person video games\cite{lample2016playing},
among others.
Recent attempts in the model-free RL domain attempt to take the existing,
discrete action space algorithms,
and improve factors such as exploration with sparse rewards~\cite{Pathak2017}\cite{Andrychowicz2017}.
Others
adapt model-free RL to a continuous action space,
with algorithms that build on Q-learning
(DDPG~\cite{lillicrap2015continuous}, TD3~\cite{Fujimoto2018})
or policy gradient methods (PPO~\cite{Henderson2017}).

In the world of robotics,
RL has been used to control robotic arms
in the context of object grasping and
movement~\cite{Zeng2018}~\cite{Singh2019}~\cite{qtopt2018}.
Many recent approaches have focused on using a continuous action-space
algorithm~\cite{Richter2019}~\cite{Singh2019}.
These methods handle continuous action spaces that more
closely resemble those of the real world.
However, these methods also tend to come with costs such as
instability~\cite{Fujimoto2018}, especially when learning from
batch data~\cite{Fujimoto2019}.

\subsection{Surgical Robotics}

Surgical robotics and needle manipulation has been studied quite heavily
recently~\cite{dettorre_2018}~\cite{sen_2016}~\cite{zhong_2019}.
Most of these works approach the problem using classical methods of path-planning and needle segmentation.
They require extensive problem and environment-specific work.
RL offers a more generalizable approach to surgical needle manipulation.

Recently, RL has started being applied to surgical environments.
\cite{Richter2019}~developed an open-source simulation based on V-REP,
which they termed dVRL, to teach a da Vinci arm
to reach a target and to move an object to a target.
It appears that only minimal physics are computed in some areas.
For example, objects are considered grabbed when they
move within a certain distance of the jaw.
\cite{Richter2019}~also focuses purely on state information,
making it limited to situations where a full state of the world can be obtained.

\cite{Varier2020}~applied Q-learning directly to a real world da Vinci robot.
They encountered the expected problems of iteration in a real world environment,
and used multiple approaches,
including limiting the robot to a grid of 1331 points.

\subsection{Batch and Off-Policy learning}

RL algorithms are fundamentally designed to work online.
As the agent samples new data,
it adjusts its policy to take account of the new data,
which then drives further agent behavior in a feedback loop.
Due to the number of steps required for RL,
the ability to collect data and run the algorithm offline is highly desirable.
Note that some batch learning is built into
the original Deep Q Network(DQN) algorithm~\cite{mnih2013playing} in the form of the replay buffer.
In theory, this algorithm family is supposed to support off-policy learning.

The practice of switching between batch and online data has been described
as early as~\cite{Lange2012}.
Recently, the ability to learn from off-policy data has been
challenged~\cite{Fujimoto2019}.
Specifically, distribution mismatch between batch data and live data can
cause the agent to assign the wrong Q value to unvisited transitions.
This makes it difficult to learn without adjusting the DQN algorithm,
as in the batch-constrained DQN~\cite{Fujimoto2019}
and distribution-constrained backup~\cite{Berkeley2019}.
\cite{Agarwal2019}~found that algorithm choice was a significant factor
when trying to learn from offline data,
with DQN doing fairly well when trained with both good and plentiful offline data.
\cite{Agarwal2019}'s recommended discrete algorithm is somewhat similar
to the more stabilized version of Double Q-Learning(DDQN) we apply here.
While we find full batch training to be limited in its training capacity
(see section~\ref{section:playback}),
using a hybrid-batch approach appears to solve this issue.

Concurrent to this work,~\cite{Nair2020} applied a similar approach to
learning using online data while also constraining using offline
data.
While we didn't have access to this approach, our approach is simpler to implement.

\section{Problem Formulation}

Q-Learning formulates RL in terms of a Markov decision process.
At state $S_t$,
an agent carrying out actions $a_t$ according to a chosen policy $\pi$,
carries out the action and obtains a reward $R_t$.

Our goal is to obtain the optimal policy $\pi^*$ by maximizing all rewards.
The value of each state is defined as the sum of all rewards including future rewards:

\begin{equation*}
  V(s_t) = \sum^{\infty}_{i=t} \gamma R_{a_i}(s_i)
\end{equation*}

In Q-learning, we focus on the value of each state-action combination.
We utilize the Bellman equations to obtain the optimal $Q^*$ value for the optimal
policy $\pi^*$.

\begin{equation*}
  Q_{target}(s_t, a_t) = {a_{t}}(s_t) + \gamma Q(s_{t+1}, \argmax_{a'}{Q(s_{t+1},a')})
\end{equation*}

where $a'$ is the set of all available actions at time $t+1$.
We incrementally adjust the current values of Q to slowly converge to $Q^*$
by minimizing the temporal difference error:

\begin{equation*}
  \delta_t = |Q(s_t, a_t) - Q_{target}(s_t, a_t)|
\end{equation*}

\subsection{RL Algorithm}

Control-based RL applications often prefer to employ continuous-action
space algorithms,
such as those that derive from Q-learning~\cite{lillicrap2015continuous}~\cite{Fujimoto2018},
and those that are policy iteration-based~\cite{Schulman2017}
These algorithms have been shown
to have issues with batch learning~\cite{Fujimoto2019}.
Additionally, we are inspired by~\cite{Zeng2018}~and~\cite{Hundt2019},
who illustrated that RL could be taught efficiently through simplification of the task.
We therefore choose instead to use the more traditional
DDQN algorithm~\cite{VanHasselt2015},
which itself improves over the original DQN algorithm~\cite{mnih2013playing}.
We further modify DDQN by adding a second set of Q-value networks
and assigning data uniquely to a Q-value network.
This moves our algorithm closer to
the original Double Q-Learning~\cite{Hasselt2010},
much as TD3 adds a secondary
neural network~\cite{Fujimoto2018}.

\subsection{State and Action Space}
\label{sec:state_rep}

We run the dVSS-RL in several modes: \emph{state} mode, where we provide
the Cartesian coordinates of the PSMs w.r.t. the camera, and some task information;
\emph{image} mode, a $224 \times 224$ pixel representation of the view
from the endoscope which may be augmented by stereo or depth data,
and \emph{mixed} mode, which combines image and minimal state data,
such as PSM coordinates.

We use velocity space with 6 DOF to provide a natural representation for the neural network.
To reduce the curse of dimensionality,
we remove unused dimensions.
For the \emph{reach task},
we reduce the action space by including only translation in 3 dimensions,
requiring a minimum of 27 possible discrete actions.
For the \emph{suture task},
we use one more angular movement (4 dimensions),
resulting in 81 possible discrete actions.
A smaller action space naturally results in faster convergence rates,
since the resulting state-action space is smaller~\cite{Varier2020}.

\subsection{Exploration and Rewards}

We leverage our access to the internals of the simulation to
shape a dense reward schedule.
The reward density allows us to minimize reliance on random exploration.

For our stochastic process,
we do not use epsilon-greedy,
but rather use a quantized Ornstein-Uhlenbeck process~\cite{uhlenbeck1930theory},
which utilizes some state.
This process seems to provide better exploration over time.

\subsubsection{Reach Task}
    This task starts with the needle being held by the right PSM
    at a constant position $P_0$.
    The agent needs to move the needle-tip as close as possible to location $T_{in}$,
    which is sampled uniformly from surface of the tissue.

    Given needle location $N$, we structure our dense reward space as follows:

    \begin{equation} \label{eqn:reach_reward}
      R = \alpha \Delta d(N,T_{in}) - \beta N_{out} - \gamma (N_{dropped} + N_{dmg}) - \eta
    \end{equation}

    where $N$ is the needle-tip location;
    $N_{out}$, $N_{dropped}$ and $N_{dmg}$
    are 1 when the needle is out of view, dropped, or penetrates the tissue, respectively,
    and 0 otherwise;
    $\alpha$ is a scaling factor we set experimentally to 10;
    $\beta$, which we set experimentally to 10, is a penalty for fundamental errors we wish to fully discourage;
    $\gamma$, which we have set to 5,
    is the penalty for \emph{incidental errors} we wish to avoid without discouraging nearby exploration;
    and $\eta$, set to 0.001, is a time penalty per step used to avoid circular motion.

\subsubsection{Suture Task}
    In this task, $T_{in}$ is the expected location of
    needle-tip ingress, and $T_{out}$ is that of needle-tip egress.
    They are separated by 8mm, which is the curvature of the needle.
    After reset, the needle is suspended directly above
    the target at set location $T_{in}$.
    The PSM needs to learn to move the needle-tip down into the ingress target,
    and out through the egress target.

    \begin{figure}[tbh]%
      \centering

      \setlength{\fboxsep}{0pt}%
      \setlength{\fboxrule}{1pt}%

      \includegraphics[width=0.9\linewidth,trim=0 20 0 50,clip]{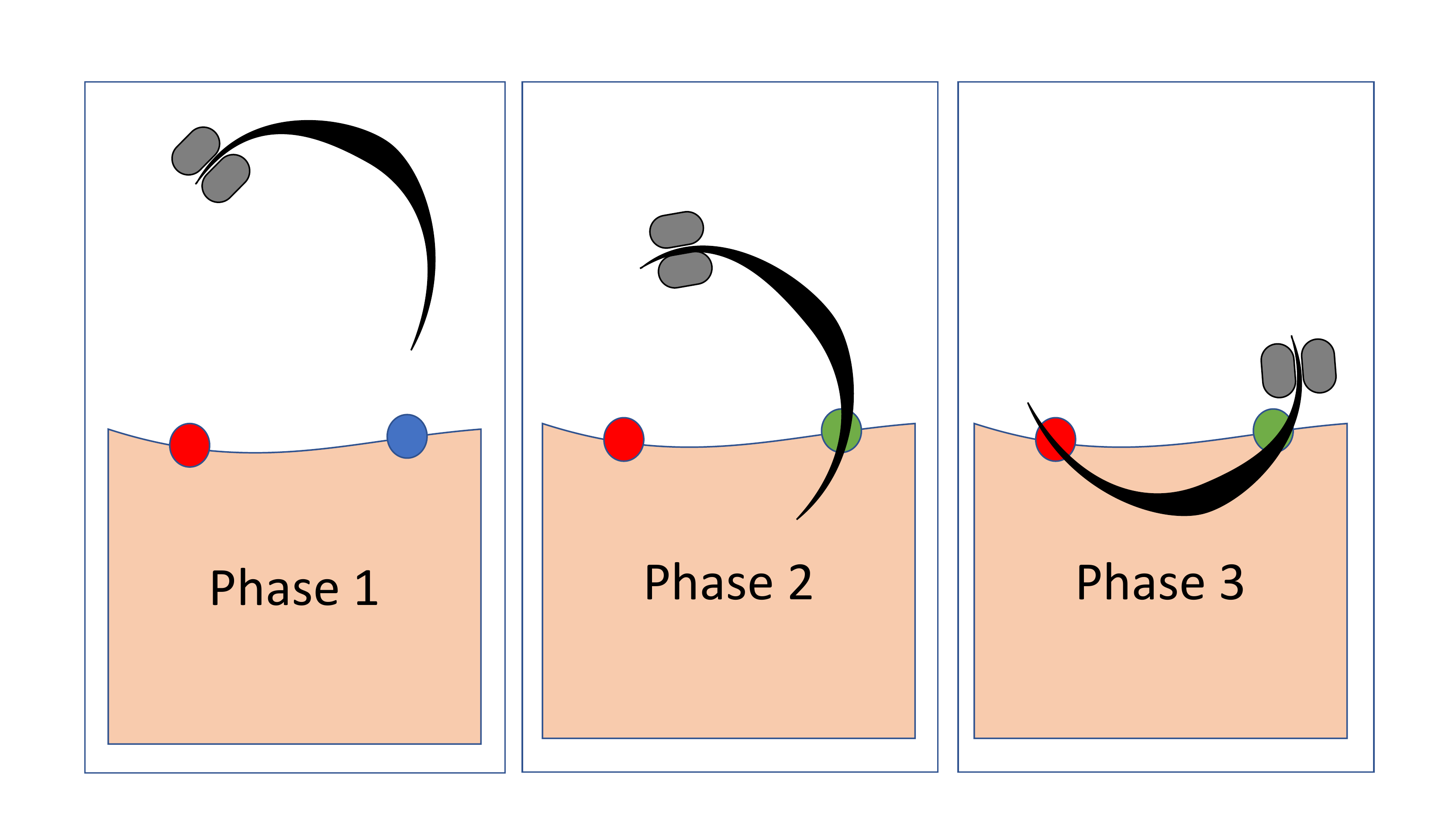}

    \caption{%
      The three phases of the \emph{suture} task.
      In phase 1, the needle remains outside of the tissue.
      Phase 2 is when the needle penetrates in ingress point.
      Phase 3 occurs when the needle exits at the egress point.
      }
      \label{fig:suture_phases}
    \end{figure}

    We differentiate between 3 separate phases of this task
    (see Figure~\ref{fig:suture_phases}).
    The reward is as follows:

    \begin{equation} \label{eqn:suture_phase2}
    \begin{split}
      R_{P1} &= \alpha \Delta d(N,T_{in}) \\
      R_{P2} &= \alpha \Delta d(N_{tip},T_{in}) + (1-\alpha) \Delta d(N_{end}, T_{out}) \\
      R &= \phi P_{next} - \psi P_{back} + P_1 R_{P1} + P_2 R_{P2} \\
        &- \beta (N_{out} + N_{dmg}) - \gamma N_{dropped} - \eta
    \end{split}
    \end{equation}

    At each phase transition, we reward the agent for advancing
    to the next stage ($P_{next}$), and penalize the agent for regressing ($P_{back}$).
    In addition, we terminate the episode on phase regression.
    For phase 1, the needle is outside the tissue,
    and we use a distance-based reward as in the reach task.
    For phase 2, when the needle enters the tissue,
    we wish to nudge the agent along the correct path.
    We do this by using the weighted sum, controlled by $\alpha$ (set to 0.8),
    of the delta distances from the needle-end
    to the ingress point and the needle-tip to the egress point.
    $P_1$, $P_2$, $P_{next}$, $P_{last}$, and $N_{dmg}$ are 1 when in phase 1; phase 2;
    advancing a phase; regressing a phase; and the needle moving laterally (causing tissue damage),
    respectively, and 0 otherwise.
    The other variables match their reach task interpretations.
    $\phi$ and $\psi$ are both set to 10,
    emphasizing the importance of task advancement and completion.

    Phase 3 occurs when the needle-tip successfully exits via the egress
    target, ending the task.

\subsection{Neural Networks}

\begin{figure*}[htb]%
  \centering

  \subfloat[Image-based Neural Network\label{fig:nn_img}]{\includegraphics[width=0.5\linewidth,trim=0 0 0 30,clip]{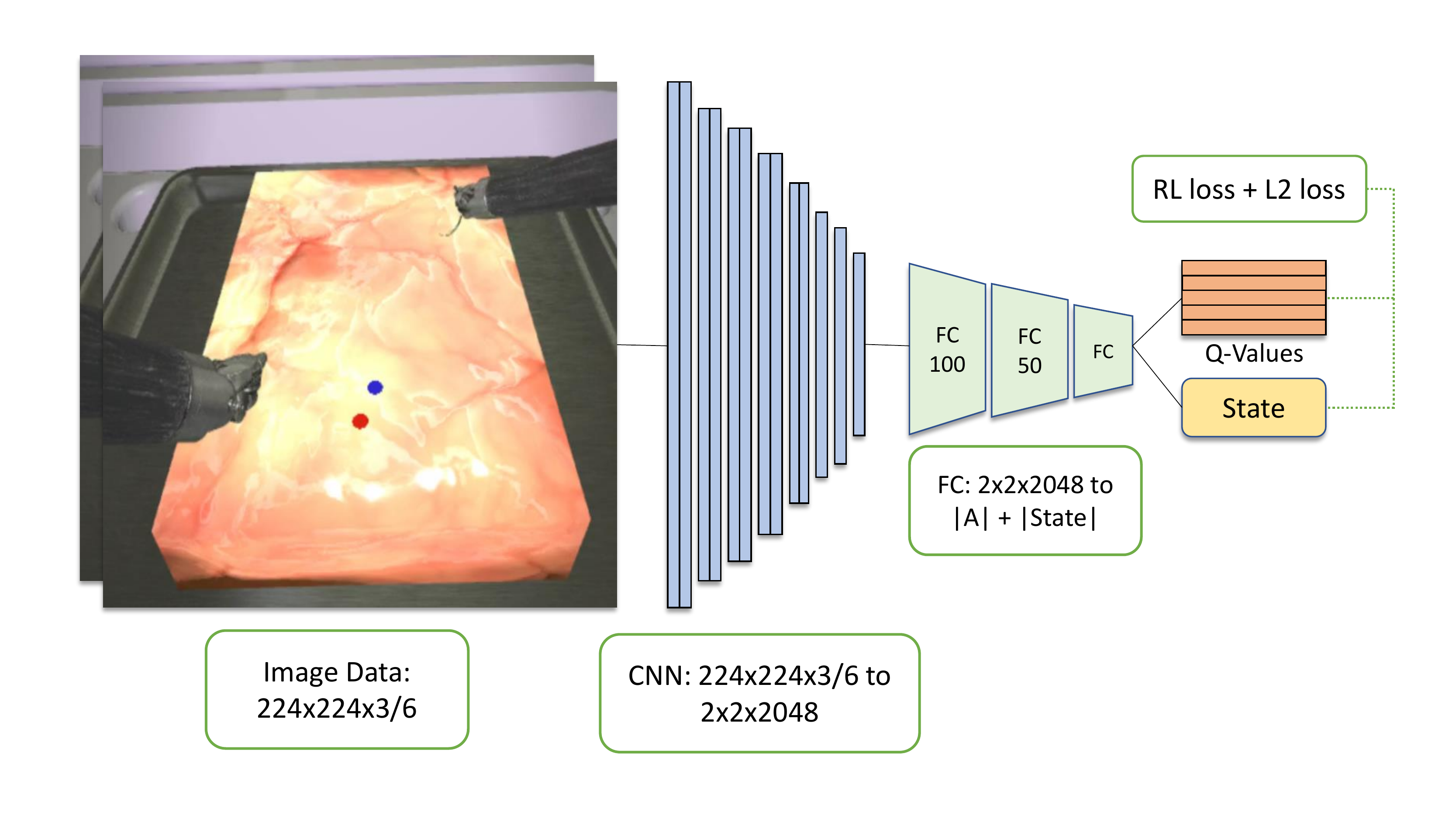}}
  \hspace{-0.02\linewidth}
  \subfloat[Mixed Neural Network\label{fig:nn_mixed}]{\includegraphics[width=0.5\linewidth,trim=0 0 0 30,clip]{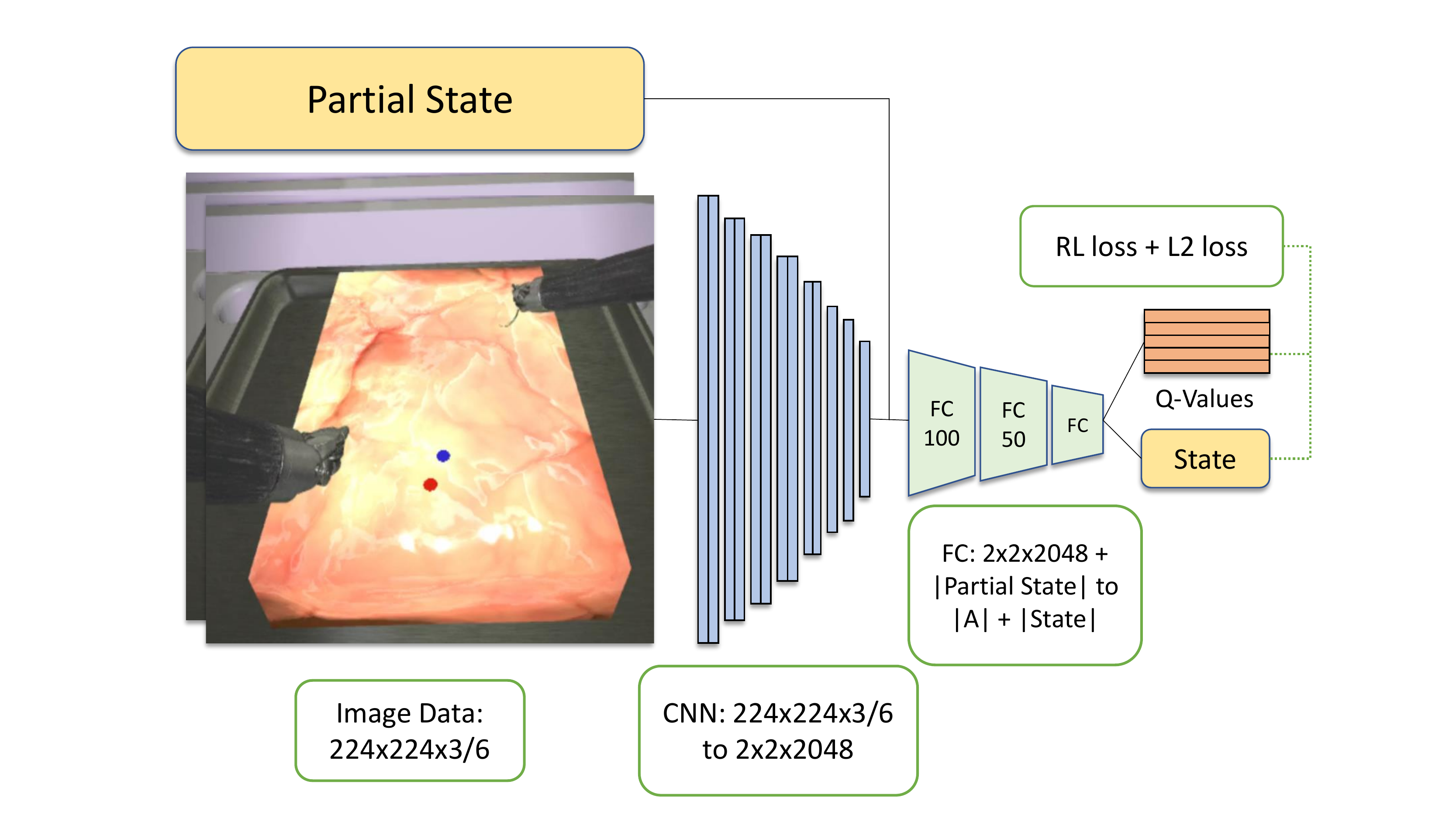}}

\caption{%
  The Neural Networks used for the image-based RL.
  Image data (mono/stereo/mono + depth) is
  fed to the CNN.
  3 fully connected layers follow.
  The loss used is a sum of the RL loss and
  auxiliary $L_2$ loss compared to the true state.
  }
  \label{fig:network}

\end{figure*}

We use a different neural network for each mode of operation.
For all networks, we use ReLU and batch normalization layers.
In \emph{state} mode, our neural network takes as input
the state specified in section~\ref{sec:state_rep}.
We add 3 fully connected layers,
gradually reducing output to $|A|$, the cardinality of our action space.
In \emph{image} mode,
we use a CNN of 8 layers, increasing feature dimensions from
3 or 6 (for mono or stereo, respectively) to 2048.
This is followed up by the same 3 layers from the \emph{state} mode network.
We use the Q-learning loss together with an auxiliary $L_2$ loss
trying to predict the state of the environment
(see Figure~\ref{fig:nn_img}).
For \emph{mixed} mode,
we add an additional state input to the \emph{image} mode.
We still use a combined loss,
consisting of the Q-learning loss together with the auxiliary $L2$ state loss
(see Figure~\ref{fig:nn_mixed}).


\section{Method}

\subsection{The dVSS-RL Environment}

The dVSS is used by Intuitive Surgical
as a training device for the real da Vinci robot.
Intuitive Surgical provided us with the source code for the simulator.
dVSS is a more full-featured simulation of the dynamics of the da Vinci
robot
than some other recent examples, such as V-REP, used by~\cite{Richter2019}.
The simulation includes two robotic arms (PSMs),
both with Large Needle Drivers (LNDs) attached.
Using the LNDs, a needle can be grasped and manipulated.
A third arm serves as the endoscope,
allowing for a realistic, high-resolution, stereo view
similar to that seen by a surgeon.
Deformable tissue is simulated, and interacts with the needle point,
allowing for a wide variety of training tasks.
Objects in the simulation are simulated to a high fidelity
using NVIDIA PhysX to compute physical forces and velocities in real-time.

We converted the dVSS into a full RL environment
which we call the dVSS-RL.
Instead of the human operator,
the RL agent controls the actions in the simulator.
Our environment hews closely to the API used
by the OpenAI Gym~\cite{OpenAIgym},
with the main actions being:

\begin{minipage}{\linewidth}
\begin{lstlisting}[basicstyle=\small,style=customPython]
  1. state = reset()
  2. data = step(action)
     (new_state, reward, done, action) = data
\end{lstlisting}
\end{minipage}

Note that the main difference from the standard OpenAI Gym API is that the \emph{step} action
returns the action taken.
This is so that the user can retrieve the \emph{actual} action taken when
using a \emph{playback} environment (see section~\ref{sec:hybrid_batch}).

We modified the simulation to render off-screen to a memory buffer using EGL
(the Khronos Native Platform Graphics Interface).
Additionally, a shared memory area was used to transfer images at high throughput,
and the simulation was modified to allow for multiple simultaneous processes.

\subsection{Hybrid Batch Learning}
\label{sec:hybrid_batch}

\begin{figure}[thb]%
  \centering

  \includegraphics[width=\linewidth,trim=0 0 0 25,clip]{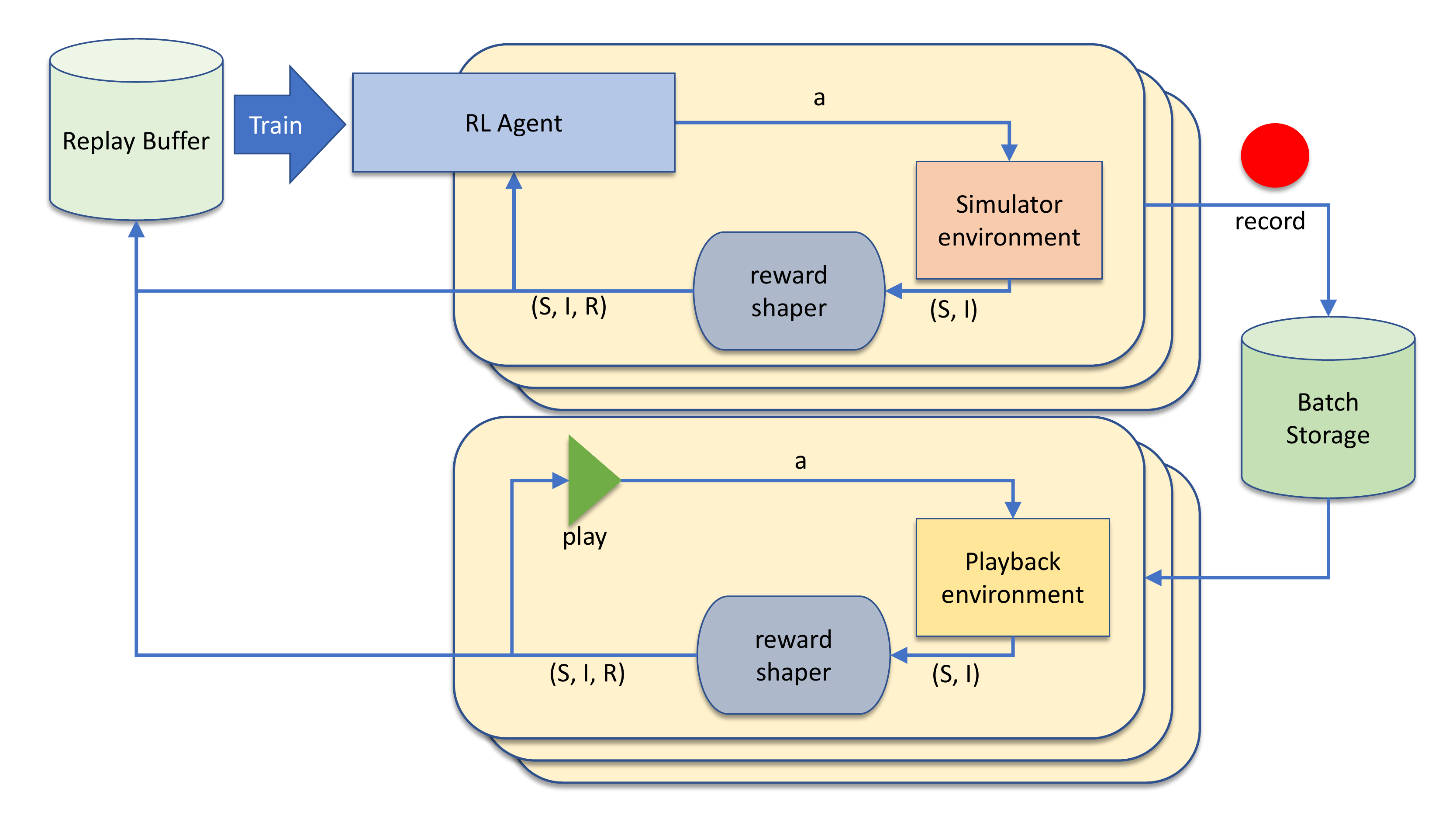}

\caption{%
  Our Hybrid Batch Learning (HBL) system.
  At the top, we see the typical RL environment loop,
  with the agent selecting an action ($a$) and feeding it into the
  simulator environment.
  The environment returns the state and image data in response.
  This is fed to the reward shaper,
  which uses state information to append a reward to the data.
  The tuple $(S,I,R)$ is saved to the replay buffer,
  which is sampled in order to train the agent.
  At the same time, all transitions are recorded in the large buffer on
  the right, and played back using playback environments.
  }
\label{fig:system}
\end{figure}

Each step of the dVSS-RL takes about 0.6 seconds,
and the performance cannot be accelerated due to the simulator's complexity.
Additionally, a reset takes around 1.5 seconds,
as the simulator needs to restore the needle position.
We therefore require a mechanism to allow us to accelerate the environment.
Realizing that this is a common need for RL systems with complex environments,
both simulated and real-world,
we make use of our Hybrid Batch Learning system (Figure~\ref{fig:system}),
which is added on top of the standard replay buffer,
to accelerate learning significantly.
While the replay buffer stores random individual transitions,
our offline batch process stores entire episodes consecutively for efficiency,
allowing us to compress
the images of each episode into a small video file.
Each step, we save both high resolution images and the portion
of simulator state required for reward computation.
When training, a mix of offline and online data is placed in the replay buffer:
we run real simulator environments that respond to agent actions,
as well as playback environments playing back old data.

HBL allows us to reuse the same off-policy data to learn
via multiple approaches.
For example, we find that learning from state is easier than
learning from images, which are high dimensional.
We can therefore apply a protocol resembling curriculum learning~\cite{Bengio2009},
exploring and training the algorithm first using state,
and then playing back mostly the same data to train using images.
Additionally, adjustments to reward shaping can be made to old data,
as the reward is not saved, but rather recomputed from state.
Hyperparameters can also be adjusted and tested more rapidly
than could be done purely on-policy.
Unlike full batch learning, however,
our method still allows the algorithm to continue to obtain useful on-policy data.

The learning time of HBL consists of RL time, sim time and playback time:

\begin{equation}
  \label{eqn:playback}
  T_{wall} = T_{RL} + N ((1 - \alpha) T_{sim} + \alpha T_{playback})
\end{equation}

where $T_{wall}$ is the total wallclock time,
$T_{RL}$ is the cost of running the RL algorithm and neural network,
$\alpha$ is the ratio of playback to sim data,
$N$ is the total number of steps,
$T_{sim}$ is the simulation time per step,
and $T_{playback}$ is the time to play back batch data per step.

For example, at $\alpha = 0.5$, with a slow enough simulation relative to playback speed,
total time can be nearly halved for the same $N$.
At $\alpha = 0.9$, we could theoretically get an order-of-magnitude improvement in performance,
assuming learning rate is not impacted.
We study this further in section~\ref{section:playback}.

While playback environment themselves are limited to previously explored state transitions,
$\alpha$ can be adjusted to enable further on-policy exploration as needed.

Note also that there is a tradeoff between measurement of agent performance and speed,
as evaluation runs must be performed on the real environment.
A dense evaluation schedule will therefore reduce HBL performance.


\section{Experimental Results}
\label{sec:results}

We run our experiments on 2 multi-core machines (AMD 3900X, Intel Xeon).
Given our desire to run environments in parallel, our bottleneck is the CPU.
For all experiments, we make maximum usage of the playback environments
and off-policy data.
We start training the easiest mode (state) and recording all data,
and then use HBL to train the more difficult modes as well,
including mono and stereo images, as well as depth + a single image.

\subsection{Reach Task}

\begin{figure}[tbh]%

  \subfloat[Reach task rewards]{\includegraphics[width=0.49\linewidth,trim=0 0 0 0,clip]{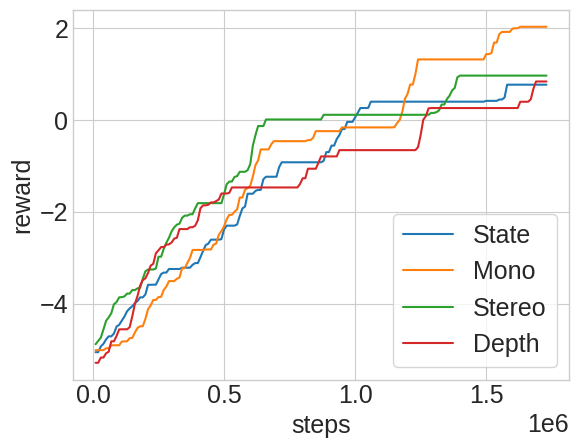}}%
  \subfloat[Reach task success]{\includegraphics[width=0.49\linewidth,trim=0 0 0 0,clip]{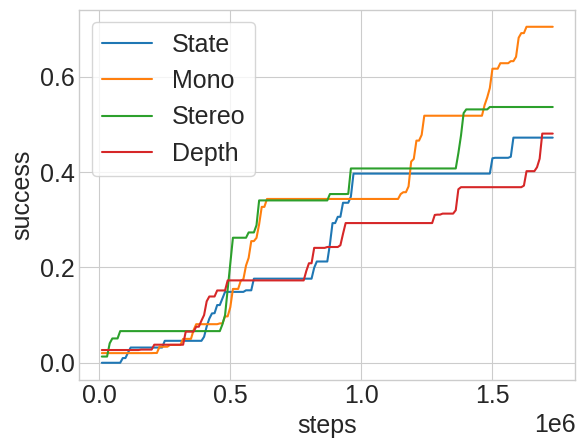}}

\caption{%
  Results for the \emph{reach} task, maximum of averages over 20,000 steps.
  }
\label{fig:reach_results}
\end{figure}

We run the \emph{reach} task with a set starting position,
and with target destinations chosen randomly over the surface of the tissue,
using 8 parallel environments.
$\alpha$ is set to 0.96.
We compare several modes:
using state alone,
using mono-ocular image data,
using stereo data,
and using one color image and one depth image.
For all modes, we use the Adam optimizer
with a learning rate of $5e^{-5}$.

Figure~\ref{fig:reach_results} shows both the maximum average reward and maximum average success
graph of this task for the different modes,
with success being defined as reaching within 2mm of the target.
Note that stereo and depth information appear to provide little additional benefit for this task.
We theorize that this could be due to the image of the needle occupying roughly 12 pixels
within the 224x224 image,
and thus providing little additional information.

\subsection{Suture Task}

\begin{figure}[tbh]%

  \subfloat[Suture task rewards]{\includegraphics[width=0.49\linewidth,trim=0 0 0 0,clip]{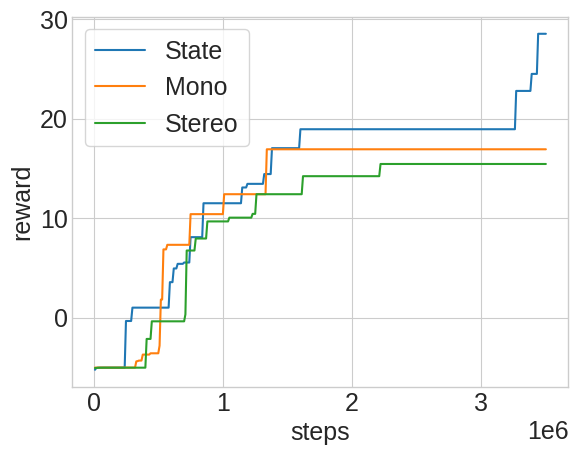}}%
  \subfloat[Suture task success]{\includegraphics[width=0.49\linewidth,trim=0 0 0 0,clip]{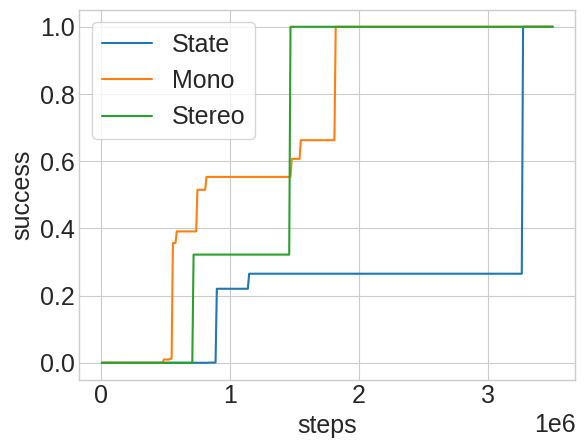}}%

\caption{%
  Maximum reward for the \emph{suture} task.
  }
\label{fig:suture_results}
\end{figure}

The \emph{suture} task simulation is far more demanding on the CPU,
as the simulator needs to compute the physical forces of the needle-tissue interaction.
We can therefore run at most 4 parallel environments.

We find that SGD with a learning rate of $1e^{-3}$ is more stable for this task.
The results can be seen in Figure~\ref{fig:suture_results}.
We plot the maximum reward and success for this task, since the layout is constant.

We compare state-based mode, mixed mono, and mixed stereo mode.
Pure image-based mode is insufficient for this task since the needle is sometimes obscured.
Note that while pure state eventually outperforms than the other modes reward-wise,
it only reaches full success later than the other modes.
We believe the additional visual data may be useful for accomplishing the task faster.

\subsection{Hybrid Batch Learning Performance}
\label{section:playback}

\begin{figure}[tbh]%
  \includegraphics[width=0.6\linewidth,trim=0 0 0 0,clip]{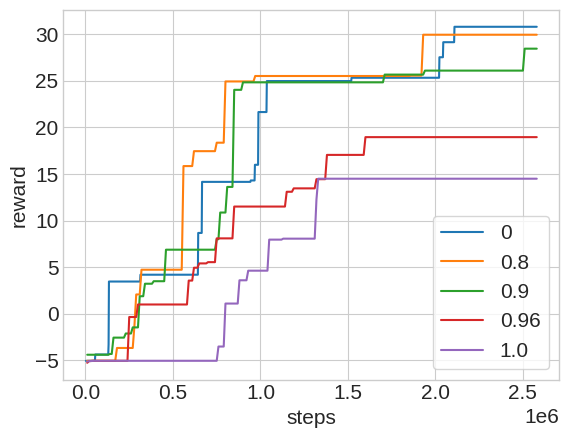}
  \caption{%
  Suture task rewards by $\alpha$.
  $\alpha$ determines how much off-policy batch data we use vs. on-policy sim data
  (see equation~\ref{eqn:playback})
  }%
  \label{fig:sut_pct_r}%
\end{figure}

\begin{table}[tbh]%
  \resizebox{0.5\linewidth}{!}{%
  \rowcolors{1}{}{lightgray}
  \begin{tabular}{ c c c }
    \hline
    $\alpha$ & Max Reward & Time (hrs)\\[1.5pt]
    \hline
    1.0 & \textbf{14.4} & 2.0 \\
    0.96  & 29.8 & \textbf{18.8} \\
    0.9  & 32.0 & 32.6 \\
    0.8  & 30.0 & 42.0 \\
    0  & 30.8 & 125.6 \\
    \hline
\end{tabular}}
  \caption{\label{tbl:sut_pct_times}Maximum reward and time taken by $\alpha$,
  the ratio of off-policy to on-policy data.%
}
\end{table}

We observe the effect of different $\alpha$ values in equation~\ref{eqn:playback}
on the \emph{suture} task.
We use SGD with a learning rate of $1e^{-3}$,
testing $\alpha$ values between 0 and 1.
To obtain lower $\alpha$ rates, we insert sleep cycles in the algorithm.

The results can be observed in Figure~\ref{fig:sut_pct_r} and table~\ref{tbl:sut_pct_times}.
We note that running at $\alpha = 1$ appears to be unreliable,
giving us rapid performance but inferior results,
perhaps due to the problems noticed by~\cite{Fujimoto2019}.
Once we mix in some live on-policy data with HBL, we obtain far better results.
It appears that going at least as high as $\alpha = 0.96$ is acceptable in this case,
resulting in 6.7 times faster wallclock performance.

\subsection{Needle Master Curriculum Tests}
\label{section:needlemaster}

\begin{figure}[htb]
  \subfloat[Agent view of NMRL]%
  {\includegraphics[width=0.35\linewidth,trim=0 0 0 0,clip]{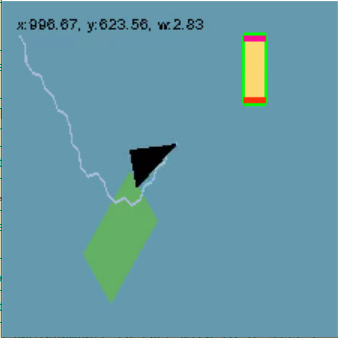}%
    \vspace{2pt}
  }
  \subfloat[Average success vs steps in NMRL]%
  {\includegraphics[width=0.55\linewidth,trim=0 0 0 0,clip]{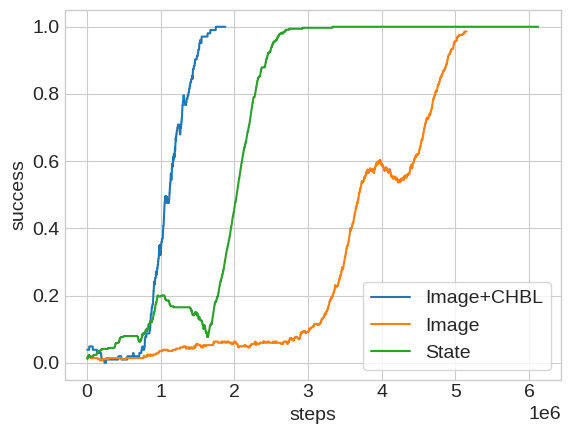}}
  \caption{On left, the agent view of Needle Master, a simplified, accelerated environment.
  The needle needs to be directed through the yellow gates in order.
  On right, average success in NMRL in image mode, with Curriculum HBL and without it,
  as compared to the original state mode used to collect data for HBL.}
  \label{fig:needle_view}
\end{figure}

To further investigate the effect of Curriculum HBL on learning performance,
we use a simpler environment where we can iterate rapidly.
Needle Master (NMRL) is a simplified 2D surgical environment we created to approximate
needle manipulation tasks.
The goal in NMRL is to learn to move the black needle through the gates via the yellow exposed areas.
NMRL supports many scenarios; we choose a fixed layout of 2 gates for our task.
See Figure~\ref{fig:needle_view}.
Success is defined as moving through both gates.
NMRL requires few CPU resources, allowing us to run 10 environments in parallel.
Each step and reset takes less than 1ms, meaning that we can rapidly get to the millions
of steps required for RL.

We find that CHBL appears to dramatically increase learning performance
\emph{regardless} of wallclock time, with respect only to steps taken.
This can be seen in Figure~\ref{fig:needle_view},
where training in image mode with CHBL takes less than \emph{half} the iterations as learning from scratch.
Image mode with CHBL also outperforms the original recording of data using state.
We theorize that this is due to the fact that the randomized state-based directed exploration
of the playback environment in CHBL is more effective than both image-based exploration
and the original state-based exploration used for recording the data.

\section{Conclusion}
\label{sec:conclusion}

We illustrate the first application of vision-based RL
to a surgical environment containing robot arms, needle and tissue.
dVSS-RL is an RL environment based on
dVSS, a high-fidelity,
realistic simulation that allows surgeons around the world to rehearse technical skills.
Despite the difficulty of training on such a complex environment in
a reasonable timeframe,
we are able to do so by using hybrid-batch learning with playback environments,
allowing us to accelerate learning by more than 6x.
Using HBL, we are also able to use state-based learning to bootstrap vision-based learning.
Our curriculum method appears to double RL effectiveness on a simple surgical RL environment.

Due to COVID-related issues, we were unable to investigate transfer to a da Vinci robot
using the dVRK.
We hope to do this in future,
and believe that the precise kinematics and dynamics of the dVSS-RL will be helpful in this endeavor.


\section*{Acknowledgements}

The authors were supported on a 2019 Intuitive Surgical Technology Grant.
We would like to thank Intuitive Surgical for access to the dVSS,
and to Cortney Jansen, Dale Bergman and Simon DiMaio for their assistance.

This work was also supported by the Office of the Assistant Secretary of Defense for Health
Affairs under Award No. W81XWH-18-1-0769. Opinions, interpretations,
conclusions and recommendations are not necessarily endorsed
by the funders.


\bibliographystyle{IEEEtran}
\bibliography{bib/drl}

\end{document}